\pdfoutput=1

\documentclass[11pt]{article}

\usepackage[]{acl}

\usepackage{times}
\usepackage{latexsym}

\usepackage[T1]{fontenc}

\usepackage[utf8]{inputenc}

\usepackage{microtype}
\usepackage{caption}
\usepackage{graphicx}
\usepackage{amsmath}
\usepackage{amssymb}
\usepackage{multirow}
\usepackage{booktabs}
\usepackage{amssymb, amsmath, bm}
\usepackage[flushleft]{threeparttable}
\usepackage{makecell, caption}
\usepackage{booktabs}

\title{Supplementary Material: Implementation and Experiments for GAU-based Model}

\author{Zhenjie Liu \\
  School of Computer Science and Technology, Xidian University \\
  {\tt santosocy18@gmail.com}}
         
\begin{document}
\maketitle
\begin{abstract}
  In February this year Google proposed a new Transformer variant called FLASH \citep{Hua:2022}, which has a faster speed, lower VRAM footprint and better performance. This is achieved by designing a performant layer named GAU (Gated Attention Unit), which combines the Attention layer and FFN. 
In this paper, some implementation details are re-analyzed both theoretically and practically.
 We then propose a novel GAU-based model and  pre-train it on a Chinese corpus. Results of the CLUE benchmark show that our model achieves a dev average score of 75.02, 1\% higher than RoFormerV1 and being 45\% faster, which is also competitive with RoFormerV2.
\end{abstract}

\section{Introduction}

These days have witnessed the great success of pre-trained Transformer-based models \citep{Vaswani:2017} \citep{Devlin:2018} \citep{Radford:2019}. The self-attention mechanism is the key defining characteristic of Transformer models, however, it's also blamed for it's quadratic time and memory complexity, which can hinder model scalability especially when processing long sequences. There has been a lot of variants proposed to address this problem by modifying the model architecture, and most of these methods fall into two categories: \textit{"sparsification"} and \textit{"linearization"}. The former \citep{Child:2019} \citep{beltagy2020longformer} \citep{zaheer2020big} introduces sparsity into attention matrix by limiting the view of each token to reduce the computation of token-to-token associativity. Furthermore, it separates the input sequence into several chunks so that the computation of attention matrix takes place in every single chunk rather than the complete sequence. To reduce the performance loss caused by reducing the view of each token, some models propose the global attention to capture the long-term dependences. But obviously, this approach has two drawbacks: 1. How to choose the area of attention to be retained is highly subjective.
2. It requires specific design optimization in programming and therefore, it's hard to generalize.

Another kind of models start by using the associative law of matrix multiplication to theoretically approximate the softmax function in the attention matrix. Some works \citep{choromanski2020rethinking} \citep{kasai2021finetuning} \citep{qin2022cosformer} design effective unbiased estimation of the original softmax with linear space and time complexity. Another models \citep{wang2020linformer} construct the approximate matrix to reduce the complexity by utilizing the low rank property of attention matrix.

Recently, Google proposed a new model architecture to address
the quality and empirical speed issues of existing Transformer variants. This is achieved by combining the Attention layer and FFN into a single unit called GAU while reducing it to just one head \citep{Hua:2022}.
However, it is flawed in many details such as the scaling factor and the replacement of softmax. In this work, we analyze several questionable details both theoretically and practically and reorganize the model architecture. In addition, we pre-train the new model on a Chinese corpus and compare it with several classical models on the CLUE benchmark \citep{xu-etal-2020-clue}. Results show that the proposed GAU-based model achieves a dev average score of 75.02, 1\% higher than RoFormerV1 and being 45\% faster. We also compared GAU and RoFormerV2 \citep{roformerv2} which both use the same hyperparameters and pre-train for the same number of steps. The comparison results show the former is slightly higher, which indicates that GAU is not inferior to RoFormerV2.

\begin{figure}[!t]
\centering
\includegraphics[scale=0.38]{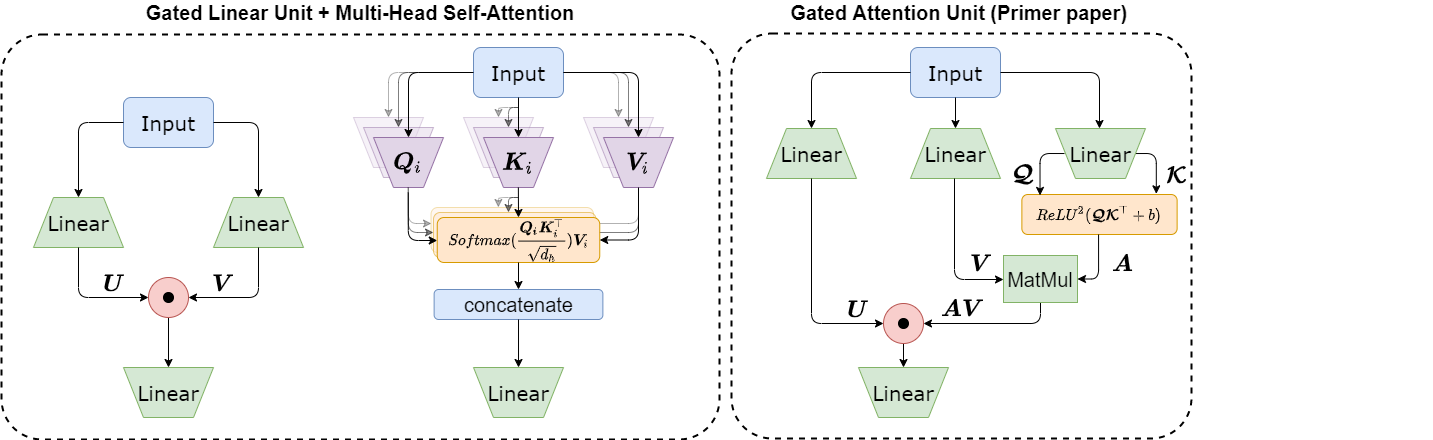}
\caption{\citep{Hua:2022} (1) The Gated Linear Unit (GLU) and the Multi-Head
Self-Attention (MHSA) unit, (2) The Gated Attention Unit (GAU) proposed in the origial paper. Ignore scaling factors and normalization  in (1), (2) for brevity.}
\label{fig:1}
\end{figure}

To summarize, our contributions include:

\begin{itemize}
		\item[$\bullet$] Many details of the original paper such as the scaling factor and the replacement of softmax are analyzed both theoretically and empirically. 
		\item[$\bullet$] We reconstruct the model architecture and pre-train it on a Chinese corpus. In addition, we also compared the fitting ability of GAU and RoFormerV2 on 9 Chinese datasets.
\end{itemize}

\section{Background and related work}
\subsection{Backbone Network: Transformer}
Given a sequence of input tokens $\{s_i\}_{i=1}^n$, the vector representations $\boldsymbol{X}\in \mathbb{R}^{n\times d_h}$ are computed via summing the word or token embedding, position and segment embedding.

\vspace*{1\baselineskip}
{\bf Multi-Head Self-Attention}. In each layer, Transformer \citep{Vaswani:2017} uses multi-head self-attention to aggregate the output vectors of the previous layer and to encode contextual information for input tokens. The operation for a single head is defined as:
$$
\boldsymbol{A}_i=softmax(\frac{\boldsymbol{Q}_i\boldsymbol{K}^{\top}_i}{\sqrt{d_h}})\boldsymbol{V}_i
$$

where $\boldsymbol{Q}_i =\boldsymbol{X}\boldsymbol{W}_q,\ \boldsymbol{K}_i =\boldsymbol{X}\boldsymbol{W}_k,\ \boldsymbol{V}_i =\boldsymbol{X}\boldsymbol{W}_v$ are obtained by applying linear transformations on the temporal dimension of the input sequence. $\boldsymbol{W}_q,\ \boldsymbol{W}_k,\ \boldsymbol{W}_v \in \mathbb{R}^{d_h\times \frac{d_h}{H}}$ are the weight matrices (learnable parameters).

{\bf Vanilla FFN}. The output for Transformer’s FFN can be formulated as follows:
$$
\begin{aligned}
\boldsymbol{X}^*=\phi(\boldsymbol{X}_A\boldsymbol{W}^{\top}_u)&\boldsymbol{W}_o\\
\boldsymbol{O}=\text{LayerNorm}(\boldsymbol{X}_A&+\boldsymbol{X}^*)
\end{aligned}
$$

\vspace*{13\baselineskip} 
where $\boldsymbol{W}^{\top}_u,\ \boldsymbol{W}_o \in \mathbb{R}^{d_{ff}\times d_h}$. Here $d_h$ denotes the hidden size of the model, $d_{ff}$ denotes the intermediate size, and $\phi$ is an element-wise activation function.

These two layers carry different functions: the Self-Attention layer is responsible for capturing the relationship between tokens, and the FFN layer can enhance the nonlinearity of the model.

\subsection{GLU}
The Gated Linear Unit \citep{Shazeer} is an improved MLP
variant augmented with gating. It has
been proven effective in many cases \citep{narang2021transformer} and is used in many state-of-the-art models. \citep{du2021glam}
$$
\begin{aligned}
&\boldsymbol{U}=\phi_u(\boldsymbol{X}\boldsymbol{W}_u),\ \boldsymbol{V}=\phi_v(\boldsymbol{X}\boldsymbol{W}_v) &\in \mathbb{R}^{n\times d_{ff}}\\
&\boldsymbol{O}=(\boldsymbol{U}\odot\boldsymbol{V})\boldsymbol{W}_o  &\in \mathbb{R}^{n\times d_{h}}
\end{aligned}
$$
where $\phi_u=\phi_v=\text{Swish}$ \citep{Swish} \citep{gelus}, and $\odot$ stands for the element-wise multiplication (Hadamard product). In GLU,
each representation $u_i$ is gated by another linear representation $v_i$ from the same token.

\subsection{GAU proposed by the original paper}
Since the GLU is more efficient FFN, we use it as the basis for modification. Notice that GLU cannot replace the Self-Attention because it lacks token-to-token interactions. To alleviate this, a natural idea is to multiply the gate value matrix $\boldsymbol{V}$ by attention matrix $\boldsymbol{A}\in\mathbb{R}^{n\times n}$:
$$
\boldsymbol{O}=(\boldsymbol{U}\odot\boldsymbol{A}\boldsymbol{V})\boldsymbol{W}_o
$$
Unlike the vanilla Transformer, the attention matrix here is calculated using the following formula:
$$
\begin{aligned}
\boldsymbol{Z}&=\phi_z(\boldsymbol{X}\boldsymbol{W}_z) &\in \mathbb{R}^{n\times s}\\
\boldsymbol{A}&=\text{ReLU}^2\left(\frac{\mathcal{Q}(\boldsymbol{Z})\mathcal{K}(\boldsymbol{Z})^{\top}}{n}+b\right)&\in \mathbb{R}^{n\times n}\\
\end{aligned}
$$
where $\boldsymbol{Z}$ is a shared linear representation of the input $\boldsymbol{X}$, $s=128$, $\mathcal{Q}$ and $\mathcal{K}$
are two simple affine transformations that apply per-dim scalars and offsets to $\boldsymbol{Z}$, and $b$ is the relative position bias. 

Compared to the standard Scaled-Dot Self-Attention, the attention matrix here is still has the complexity of $O(n^2)$. However, it has been improved in many ways to improve computational efficiency and reduce the number of parameters. The biggest change is to replace the softmax function with squared ReLU which is obtained by NAS method \citep{so2021primer}. This replacement also shows that the softmax in the attention mechanism is not necessary and can be replaced by a regular activation function with a simple normalization method. Another significant improvement is that the GAU uses only one head, which greatly enhances computing efficiency and reduces VRAM usage.

The Transformer’s MHSA comes with $4d_h^2$ parameters, and the FFN layer has $2d_h\times d_{ff}$ parameters. In standard Transformer-based models like BERT, the $d_{ff}$ is set to be $4d_h$. However, the GAU block here only has $3d_h\times d_{ff}$ parameters. ($\boldsymbol{W}_z$, scalars and
offsets in Q and K are negligible) By setting $d_{ff} = 2d_h$ for GAU, this compact design allows us to replace each Transformer Encoder block (MHSA + FFN) with two GAU layers while retaining similar training speed and fitting ability.

\subsection{RoPE}
Rotary Position Embedding (RoPE) \citep{su2021roformer} encodes absolute position information with rotation matrix. Compared to sinusoidal position embedding proposed in Transformer, the later is additive, while the former can be considered multiplicative.

\begin{table}
\centering
\begin{tabular}{cc}
\toprule[1.5pt]
\textbf{Scaling Factors} & \textbf{MLM Acc}\\
\toprule[1.5pt]
$n^2$ & 20.16\% \\
$n$ & 19.34\% \\
$n\cdot s$ & \textbf{23.11}\% \\
$s^2$ & misconvergence \\
\bottomrule[1.5pt]
\end{tabular}
\caption{MLM results of various scaling factors.}
\label{tab:1}
\end{table}

We can add absolute position information to $\boldsymbol{q}$ and $\boldsymbol{k}$ which are the linear representations of the input using the following formula:
$$
\begin{aligned}
\tilde{\boldsymbol{q}}_m = \boldsymbol{f}(\boldsymbol{q}, m)=&\begin{pmatrix}q_0 \\ q_1 \\ q_2 \\ q_3 \\ \vdots \\ q_{d-2} \\ q_{d-1} 
\end{pmatrix}\odot\begin{pmatrix}\cos m\theta_0 \\ \cos m\theta_0 \\ \cos m\theta_1 \\ \cos m\theta_1 \\ \vdots \\ \cos m\theta_{d/2-1} \\ \cos m\theta_{d/2-1} 
\end{pmatrix} + \\&
\begin{pmatrix}-q_1 \\ q_0 \\ -q_3 \\ q_2 \\ \vdots \\ -q_{d-1} \\ q_{d-2} 
\end{pmatrix}\odot\begin{pmatrix}\sin m\theta_0 \\ \sin m\theta_0 \\ \sin m\theta_1 \\ \sin m\theta_1 \\ \vdots \\ \sin m\theta_{d/2-1} \\ \sin m\theta_{d/2-1} 
\end{pmatrix}
\end{aligned}
$$
Thus, after the inner product operation, the result attention matrix will carries relative position information.

\section{Details}
\label{sec:details}
In this section we will discuss several details in the original paper. We first analyse whether to use dropout in section~\ref{sec:dropout}, we then investigate which scaling factor is best in ~\ref{sec:scaling}. And finally, we compare squared relu function with softmax in terms of the fitting and the generalization ability in section~\ref{sec:compare}.

\subsection{Dropout}
\label{sec:dropout}
In the original paper, there is no discussion of which part of the model should have dropout added. And furthermore, the dropout rate is set to 0 in appendix section B. Considering that the original FLASH model was pre-trained and tested in processing long sequences, overfitting is not a major constraint on it. For the sake of computational efficiency, dropout was removed. While the maximum sequence length of our model is set to 512, we thus introduce dropout in the linear and attention layers like other prevalent models.

\begin{table*}[htb!]
\caption{MLM accuracy and CLUE dev average score of different activation functions and training strategies. }
  \begin{center}
  \begin{threeparttable}
  \footnotesize
    \renewcommand{\TPTminimum}{\linewidth}
    \makebox[\linewidth]{%
    \tabcolsep=0.11cm
    \setlength{\extrarowheight}{2pt}
    \begin{tabular}{cccccc} 
\toprule[1.5pt]
\multirow{2}{*}{\textbf{Model}} & \multicolumn{2}{c}{\textbf{Training Strategy}\tnote{4}} & \multirow{2}{*}{\textbf{MLM Acc}} & \multirow{2}{*}{\textbf{CLUE}} & \multirow{2}{*}{\textbf{Average Pre-training Time}} \\ 
\cline{2-3}& \multicolumn{1}{l}{MLM} & 
\textbf{Fine-tune}&&\\ \toprule[1.5pt]
\multirow{2}{*}{$\frac{1}{ns}\text{ReLU}^2(\frac{1}{\sqrt{d_h}})$}& \multirow{1}{*}{$512$}&\multirow{2}{*}{$diff$}& 25.23\% &42.37  & \multirow{2}{*}{07:32:34}\\
& $diff$  &  & 24.01\%   & 50.88   \\\hline
\multirow{2}{*}{$\frac{1}{c_i \cdot ns}\text{ReLU}^2(\frac{1}{\sqrt{d_h}})$\tnote{1}}& \multirow{1}{*}{$512$}&\multirow{2}{*}{$diff$}& 31.97\% &55.41  & \multirow{2}{*}{08:22:51}\\
& $diff$  &  & 30.18\%  & 58.94   \\\hline
\multirow{2}{*}{$softmax(\frac{1}{\sqrt{d_h}})$\tnote{2}}& \multirow{1}{*}{$512$}&\multirow{2}{*}{$diff$}& 40.86\% &62.8  & \multirow{2}{*}{\textbf{07:11:13}}\\
& $diff$  &  & \textbf{41.2}\%   & 65.49   \\\hline
\multirow{2}{*}{$softmax(\frac{\log_{512}n}{\sqrt{d_h}})$\tnote{3}}& \multirow{1}{*}{$512$}&\multirow{2}{*}{$diff$}& 41.07\% &64.23  & \multirow{2}{*}{07:31:22}\\
& $diff$  &  & 40.62\%   & \textbf{66.52}   \\
\bottomrule[1.5pt]
\end{tabular}}
\end{threeparttable}
  \end{center}
      \begin{tablenotes}[flushleft]
       \footnotesize
       \item \small{1. The scaled squared ReLU function.$\quad$ 2. The standard softmax in attention mechanism.}
       \item \small{3. A softmax variant proposed by \citep{kexuefm-8823} which improves generalization ability. Hereinafter called $softmax\_plus$.}
       \item \small{4. Different training strategies in pre-training and fine-tuning phase. "\textit{diff}" represents that we feed the model with data of different length, and "\textit{512}" means that the input sequence length is fixed at 512. Results are obtained by pre-training for 10k steps.}
        \end{tablenotes}
\label{tab:2}
\end{table*}

\subsection{Scaling Factor}
\label{sec:scaling}
According to the reference code in the appendix of the original paper, the scaling factor without bias is $\frac{1}{n^2}$ after simplification, which is smaller than the standard Transformer of $\frac{1}{n}$.

We train the model under diffenrent scaling factors for 5k steps and report the result
in Table~\ref{tab:1}, from which we can see that $n^2$ is not an optimal choice and the model will perform better by replacing it with $n\cdot s$.

\subsection{Squared ReLU or Softmax}
\label{sec:compare}
It's well known that two most important capabilities of a NLP model is the fitting ability and the generalization ability. And we will compare squared ReLU and softmax in these two respects to determine whether using the Squared ReLU function is the best choice.

{\bf Generalization ability.} We enhance the fitting ability by pre-training on the large-scale unsupervised corpus. However, for prevalent NLP models like BERT and GPT, we do not design specific tasks to improve the generalization ability.

Let's take the input sequence length for a fairly deep dive. In the pre-training phase, the value of $n$ is nearly identical to the max sequence length. Then we directly use the pre-trained model for various tasks as if the model can automatically generalize to different input seqence lengths. However, results in the Table~\ref{tab:2} show that when we replace the softmax function with squared ReLU, results of downstream tasks with variable-length inputs turn out badly. This means that compared with softmax, the squared ReLU function is much worse in the generalization ability over the input sequence length.

In the softmax formula, the only thing associated with the length $n$ is the scaling factor $c_i$: 
$$
\begin{aligned}
a_{i,j} = \frac{1}{c_i}\exp\left(\frac{\boldsymbol{q}_i\cdot\boldsymbol{k}_j}{\sqrt{d_h}}\right), c_i = \sum_{i=1}^n \exp\left(\frac{\boldsymbol{q}_i\cdot\boldsymbol{k}_j}{\sqrt{d_h}}\right)
\end{aligned}
$$
To investigate the effect of scaling factor on generalization ability, we conducted some experiments. Table~\ref{tab:2} shows that the scaled squared ReLU: 
$$\begin{aligned}
a_{i,j} = \frac{1}{c_i\cdot ns}\text{ReLU}^2\left(\frac{\boldsymbol{q}_i\cdot\boldsymbol{k}_j}{\sqrt{d_h}}\right) \\
c_i = \sum_{i=1}^n \text{ReLU}^2\left(\frac{\boldsymbol{q}_i\cdot\boldsymbol{k}_j}{\sqrt{d_h}}\right)
\end{aligned}
$$
has a good performance on length generalization.

Besides adding a normalization coefficient to the original function, \citep{kexuefm-8823} proposed a novel softmax variant in terms of entropy invariance which is expressed as:
$$
\boldsymbol{A} = softmax\left(\frac{\log_{512} n}{\sqrt{d_h}}\boldsymbol{Q}\boldsymbol{K}^{\top}\right)V
$$
$a_{ij}$ can be considered as a conditional distribution of $j$ given $i$. The entropy of it can be calculated as:
$$
h_i = -\sum_{j=1}^n a_{i,j}\log a_{i,j}
$$
Assume that $h_i$ represents the degree of focus of the $\text{i}^{\text{th}}$ token on each token. Specially, if $\boldsymbol{a}_i$ is an uniform distribution, it's entropy is:
$$
a_{i,j}=\frac{1}{n},h_{i}=-\sum_{i=1}^n a_{i,j}\log{a_{i,j}}=\log{n}
$$
And if $h_i \propto n$, then $\log{h_i} \propto \log{n}$. To some extent, it indicates that the attention mechanism here degenerates into an uniform distribution, which cannot fully capture the token-to-token interactions. Therefore, $h_i$ should be length-insensitive.

Intuitively speaking, we want the distribution of attention of the $\text{i}^{\text{th}}$ token remains the same rather than being disturbed after the introduction of new tokens. \citep{kexuefm-8823} analyzed this problem theoretically and proposed a new variant called $softmax\_plus$. And results in Table~\ref{tab:2} show that it's beneficial to improve the generalization over the input sequence length by adding a scaling factor inside the softmax function.

From the analysis above, we have two solutions with admirable generalization performance:
\begin{itemize}
		\item Scaled Squared ReLU function;
		\item softmax\_plus function.
\end{itemize}

\begin{table}
\centering
\begin{tabular}{ccc}
\toprule[1.5pt]
\textbf{Matrix} & \textbf{Rank/max\_len} & \textbf{Sparsity} \\
\toprule[1.5pt]
$\boldsymbol{Q}\boldsymbol{K}^{\top}$ & 0.25 & $\approx$ 0 \\
$softmax(\frac{\boldsymbol{Q}\boldsymbol{K}^{\top}}{\sqrt{d_h}})$ & 0.9983 & 10.03\%\\
$\text{ReLU}^2(\frac{\boldsymbol{Q}\boldsymbol{K}^{\top}}{\sqrt{d_h}})$ & 0.989 & 12.49\% \\
\bottomrule[1.5pt]
\end{tabular}
\caption{Ratio of rank to max input length (512) of different attention matrices.}
\label{tab:3}
\end{table}

However, It is worth noting that in terms of the computational efficiency, the later runs faster than the former. And thus the softmax\_plus function might be a better choice.

{\bf Fitting ability.} The fitting ability of an NLP model largely depends on its capability to capture the interactions between tokens, which can be mathematically represented by the rank of it's attention matrix.

\begin{figure}[!t]
\centering
\includegraphics[scale=0.42]{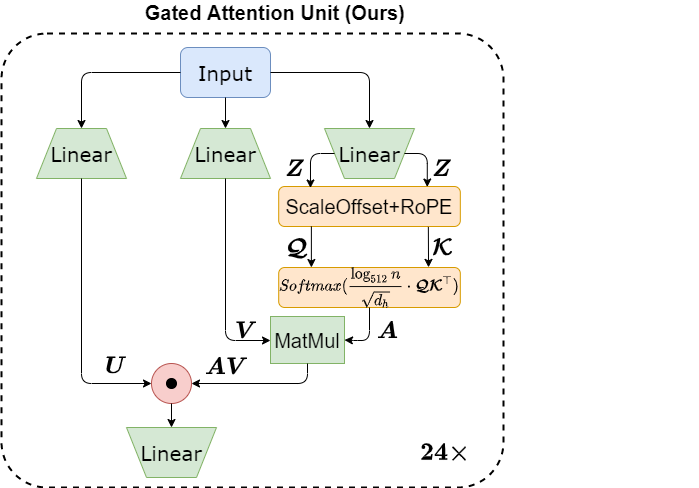}
\caption{Our proposed GAU-based model.}
\label{fig:2}
\end{figure}

Since the attention matrix without softmax is obtained by multiplying two low-rank matrices $\boldsymbol{Q}, \boldsymbol{K} \in \mathbb{R}^{n\times s}$, we have $R(\boldsymbol{Q}\boldsymbol{K}^{\top})\le s$, where $s=128$. Due to it's low rank property, we use softmax to obtain a high-rank matrix. As shown in Table~\ref{tab:3}, the attention matrix after softmax is close to be full rank. However, after applying the relu function, the rank of the attention matrix is not as high as applying softmax, which indicates that it carries less information. This is probably because the matrix is more sparse after applying the relu function. Since the relu function replaces negative values with zero, many relationships between tokens are lost.

So in terms of the fitting ability of the model, using the softmax fuction could be a better choice than replacing it with squared relu. And according all factors aforementioned, we use softmax\_plus in our reorganized model architecture.

\section{Methodology}
Combining the analysis in Section~\ref{sec:details} with some conclusions from the original paper, we propose a novel GAU-based model. Model architecture and several pre-training details are showcased in section~\ref{sec:architecture} and ~\ref{sec:pre-training}.

\subsection{Model architecture}
\label{sec:architecture}
As shown in Figure~\ref{fig:2}, our model is built upon 24-layers GAU. Let $\boldsymbol{X}\in \mathbb{R}^{n\times d_h}$ be the representations over $n$ tokens, and $\boldsymbol{Z}\in \mathbb{R}^{n\times s}$ is the shared linear representation. The attention matrix can be calculated as:
$$
\begin{aligned}
\boldsymbol{\mathcal{Q}}&=w_Q\boldsymbol{Z}+b_Q+RoPE &\in \mathbb{R}^{n\times s}\\
\boldsymbol{A}&=softmax(\frac{log_{512} n}{\sqrt{d_h}}\boldsymbol{\mathcal{Q}}\boldsymbol{\mathcal{K}}^{\top})&\in \mathbb{R}^{n\times n}\\
\end{aligned}
$$
\begin{table*}[ht]
\caption{Comparison of VRAM footprint and speedup between proposed model and baselines in pre-training phase.}
\label{tab:4}
\begin{threeparttable}
\resizebox{1.0\textwidth}{!}{
\begin{tabular}{@{}c|cccccccc|ccc}
\toprule[1.5pt]
\multirow{3}{*}{\textbf{Models}} & \multicolumn{8}{c|}{\textbf{Input Sequence Length}} &\multirow{3}{*}{\textbf{Params}}
\\ \cmidrule(l){2-9} 
 & \multicolumn{2}{c|}{\textbf{256}} & \multicolumn{2}{c|}{\textbf{512}} & \multicolumn{2}{c|}{\textbf{1024}} & \multicolumn{2}{c|}{\textbf{2048}} \\ \cmidrule(l){2-9}
     & VRAM (M)\tnote{1}& \multicolumn{1}{c|}{Time Cost\tnote{2}} 
 & VRAM (M)\tnote{1}& \multicolumn{1}{c|}{Time Cost\tnote{2}} 
 & VRAM (M)\tnote{1}& \multicolumn{1}{c|}{Time Cost\tnote{2}} 
 & VRAM (M)\tnote{1}& Time Cost\tnote{2} \\ \midrule[1.5pt]
\multicolumn{1}{l|}{RoFormerV1} & 6125 (105.95\%) & \multicolumn{1}{c|}{1.487$\times$} & 10549 (124.6\%) & \multicolumn{1}{c|}{1.45$\times$} & 25059 (169.1\%) & \multicolumn{1}{c|}{1.582$\times$} & OOM\tnote{3} &  - & 124,148k (128.63\%)  \\
\multicolumn{1}{l|}{RoFormerV2} & 5643 (97.61\%) & \multicolumn{1}{c|}{1.12$\times$} & 10047 (118.7\%) & \multicolumn{1}{c|}{1.224$\times$} & 19631 (132.5\%) & \multicolumn{1}{c|}{1.471$\times$} & 45277 (153.3\%) &  1.902$\times$ & 94,777k (98.2\%) \\
\multicolumn{1}{l|}{GAU (ours)} & 5781 (100\%) & \multicolumn{1}{c|}{1$\times$} & 8463 (100\%) & \multicolumn{1}{c|}{1$\times$} & 14819 (100\%) & \multicolumn{1}{c|}{1$\times$} & 29533 (100\%) &  1$\times$ & 96,519k (100\%) \\
\bottomrule[1.5pt]
\end{tabular}
}
\begin{tablenotes}
  \item \small{1. batch size = 8.$\quad$2. Measured based on time cost of pre-training for 1k steps.  Using a single
T40 GPU.} 
  \item \small{3. OOM stands for CUDA out of memory.}
\end{tablenotes}
\end{threeparttable}
\end{table*}

\begin{table*}[ht]
\centering
\small 
\caption{Best averaged results on the evalulation datasets of CLUE.}
\begin{threeparttable}
\begin{tabular}{c|cccccc|ccc|c}
\toprule[1.5pt]
\textbf{Model} &  \textbf{AFQMC} & \textbf{CSL}& \textbf{IFLYTEK}& \textbf{TNEWS}& \textbf{WSC}& \textbf{CMNLI}& \textbf{CMRC}& \textbf{CHID}& \textbf{C3}& \textbf{AVG}\\ \midrule[1.5pt]
RoFormerV1 & 74.21 & 83.13 & 60.17 & 58.07 & 83.22 & 81.5 & 74.31 & \textbf{86.21} & 65.27& 74.01\\
RoFormerV2 & \textbf{75.96} & \textbf{84.81} & \textbf{63.24} & \textbf{59.39} & \textbf{83.93} & 81.41 & \textbf{79.35} & 85.63 & \textbf{74.32}& \textbf{76.45}  \\
GAU (ours) & 74.51 & 83.7 & 62.72 & 57.93  & 82.89 & \textbf{81.97} & 78.04 & 85.49 & 67.98 & 75.02  \\\hline
RoFormerV2* & \textbf{70.66} & 79.13 & \textbf{59.07} & 55.38 & 63.82 & 77.26 & \textbf{70.25} & 77.08 & 53.86& 67.39\\
GAU (ours)* & 69.14 & \textbf{79.6} & 58.36 & \textbf{56.57} & \textbf{64.11} & \textbf{77.47} & 68.86 & \textbf{78.2} & \textbf{56.15}& \textbf{67.61}  \\
\bottomrule[1.5pt]
\end{tabular}
\begin{tablenotes}
  \item[*] \small{Pre-trained for 30k steps using a MLM-only approach.} 
\end{tablenotes}
\end{threeparttable}
\label{tab:5}
\end{table*}

We apply per-dim scaling and offset to $\boldsymbol{Z}$ which is very cheap. Additionally, we add RoPE to $\boldsymbol{\mathcal{Q}}$ to further enhance the generalization ability of the model. \citep{su2021roformer}

It is good to note that we apply a \textit{post-norm} \citep{1000layer} to GAU layer output:
$$
hidden\_states^{l+1} = Norm(hidden\_states^{l}+\boldsymbol{O})
$$
we replace the layer normalization with $Norm$ function where:
$$
Norm(\boldsymbol{X})=\frac{\boldsymbol{X}}{\sqrt{VAR(\boldsymbol{X})+\epsilon}}
$$

According to the analysis of \citep{kexuefm-9009}, the pre-norm structure is equivalent to increasing the width of the model and decreasing the depth, and thus has underperformed the post-norm.

\subsection{Model pre-training}
\label{sec:pre-training}
We use the CLUECorpusSmall (14G) from CLUE for WWM pre-training \citep{cui2021pre}. See
Appendix \ref{wwm} for detailed settings.

\section{Experimental results}
In this section, we will verify the effectiveness (\ref{sec:clue}) and efficiency (\ref{sec:efficiency}) of proposed model and several baseline models on various CLUE tasks with detailed explanations.

\subsection{Baselines}
\label{sec:baselines}
First of all, the RoFormerV1 \footnote{For this experiment, we adopt code (Apache-2.0 License) from https://github.com/JunnYu/RoFormer\_pytorch.} model \citep{su2021roformer} is included as a standard baseline for calibration. It proposed RoPE which is used in this paper. And to demonstrate the advantages of our proposed GAU layer, we include RoFormerV2 \citep{roformerv2} as a much stronger baseline. Compared to RoFormerV1, RoFormerV2 removes all biases and replaces layer normalization with RMS-Norm, which is consistent with our model. 

\subsection{Computational efficiency}
\label{sec:efficiency}
We list comparison results on computational efficiency of 3 models in Table~\ref{tab:4}, from which we can infer that the GAU-based model has a significant improvement in VRAM usage and training speed compared to baselines.

\subsection{Evaluation on CLUE}
\label{sec:clue}
As shown in Table~\ref{tab:5}, GAU achieves comparable overall performance to RoFormerV2 on various downstream tasks when fine-tuned. Importantly, it outperforms RoFormerV1 by an absolute improvement of 1\%.

Since Roformerv2 was pre-trained on a larger corpus using the multi-task approach, we also compared GAU with RoformerV2 MLM-only version for fair comparison. Both of them are implemented in the same codebase with identical hyper-parameters and pre-trained for 30k steps. We table their comparison results in Table~\ref{tab:5}. The metrics show the effectiveness of our proposed GAU structure.

\section{Conclusion}
In this paper, the GAU architecture in the original paper is re-analyzed from several perspectives. The discussion of the scaling factors illustrates that replacing $n^2$ with $n\cdot s$ helps to improve the model performance. Besides, we compare $\text{ReLU}^2$ and softmax both theoretically and empirically. The results show that softmax has a significant advantage over $\text{ReLU}^2$ in both length generalization ability and fitting ability. We then present a novel model based on GAU. Experiments on whole word mask language modeling task shows that it is as good as RoFormerV2. Finally, experiment on nine tasks from CLUE demonstrate the superior performance of our model when applied to downstream tasks.

\section{Discussion and Future work}
The success of GAU depends largely on GLU's efficiency. GLU explicitly introduces control over the flow of information, which is equivalent to having a priori information. However, there is a lack of rigorous theoretical proof for the superiority of GLU over standard FFN. 
A future work is to investigate this and whether GLU can replace FFN in other scenarios. In addition, we will also pre-train our model on a larger corpus and apply it on a wider range of NLP tasks.

\bibliography{custom}
\bibliographystyle{acl_natbib}
\clearpage
\appendix
\onecolumn

\section{CLUE Benchmark}
We choose 6 classification datasets (AFQMC, CMNLI, CSL, IFLYTEK, TNews, WSC) and 3 machine reading comprehension datasets (CMRC2018, CHID, C3) which are all from CLUE Benchmark \citep{xu-etal-2020-clue}. Hyperparameters for tasks above are listed in Table~\ref{tab:6}.

\begin{table*}[!h]
\centering
\small
\caption{Hyperparameters for CLUE datasets.}
\begin{threeparttable}
\begin{tabular}{@{}c|cccccc|ccc}
\toprule
 &  \textbf{AFQMC} & \textbf{CSL}& \textbf{IFLYTEK}& \textbf{TNEWS}& \textbf{WSC}& \textbf{CMNLI}& \textbf{CMRC}*& \textbf{CHID}& \textbf{C3}\\ \midrule
Sequence length & \multicolumn{9}{c}{512} \\
Batch size & 16 & 16 & 32 & 16 & 16 & 32 & 32 & 32 & 24\\
Epochs & 3 & 4 & 5 & 3 & 40 & 3 & 3 & 3 & 8  \\
Peak learning rate & 3e-5 & 2e-5 & 3e-5 & 2e-5  & 1e-5 & 3e-5 & 3e-5 & 3e-5 & 2e-5  \\
Warmup proportion & \multicolumn{9}{c}{0.1} \\
Learning rate decay & \multicolumn{9}{c}{Linear} \\
Optimizer & \multicolumn{9}{c}{AdamW} \\
Adam $\epsilon$ & \multicolumn{9}{c}{1e-8} \\
Adam $(\beta_1, \beta_2)$ & \multicolumn{9}{c}{(0.9, 0.999)} \\
Weight decay & \multicolumn{9}{c}{0.01} \\ 
Hidden dropout & \multicolumn{9}{c}{0.1} \\
Attention dropout & \multicolumn{9}{c}{0.1} \\
Classifier dropout & \multicolumn{9}{c}{0.1} \\
\bottomrule
\end{tabular}
\begin{tablenotes}
  \item[*] \small{We use the average score of F1-score and EM for CMRC dataset and accuracy for the remaining as the evaluation metrics.} 
\end{tablenotes}
\end{threeparttable}
\label{tab:6}
\end{table*}

\section{Hyperparameters for MLM pre-training}
\label{wwm}
Hyperparameters for the MLM task on CLUECorpusSmall are listed in Table~\ref{tab:7}.

\begin{table*}[!h]
\centering
\caption{Hyperparameters for the MLM task on CLUECorpusSmall}
\begin{threeparttable}
\begin{tabular}{@{}c|c@{}}
\toprule
 &  MLM Results \\ \midrule
Data & CLUECorpusSmall (14G) \\
Sequence length & $512$ or $diff$ \\
Batch size &  64\\
Gradient accumulation steps &  4\\
Number of steps & 5k, 10k, 30k, 100k \\
Warmup proportion & 0.1 \\
Peak learning rate & 3e-4  \\
Learning rate decay & Linear \\
Optimizer & AdamW \\
Adam $\epsilon$ & 1e-6 \\
Adam $(\beta_1, \beta_2)$ & (0.9, 0.999) \\
Weight decay & 0.01 \\ 
Hidden dropout & 0.1 \\
Attention dropout & 0.1 \\
Classifier dropout & 0.1 \\
\bottomrule
\end{tabular}
\end{threeparttable}
\label{tab:7}
\end{table*}

\end{document}